%% file: root.tex
\documentclass[letterpaper, 10 pt, conference]{ieeeconf}  
\IEEEoverridecommandlockouts 
\title{\LARGE \bf
STUN: Self-Teaching Uncertainty Estimation for Place Recognition
} 
\author{Kaiwen Cai\authorrefmark{1}, 
Chris Xiaoxuan Lu\authorrefmark{2}, 
Xiaowei Huang\authorrefmark{1}\\
\authorrefmark{1}University of Liverpool, \authorrefmark{2}University of Edinburgh
}

\usepackage{color}

\definecolor{RightColor}{RGB}{39, 174, 96} 
\definecolor{WrongColor}{RGB}{202, 62, 71} 
\definecolor{QueryColor}{RGB}{155, 89, 182} 

\usepackage{amsmath}
\usepackage{amsfonts}

\usepackage{dsfont}

\usepackage{bbm}

\usepackage{siunitx}

\usepackage{graphicx}
\graphicspath{ {./figures/} }

\usepackage{booktabs}
\usepackage{float}  
\usepackage{threeparttable} 
\usepackage{tablefootnote}  
\usepackage{multirow}

\usepackage{cite}
\usepackage{color}
\usepackage[dvipsnames]{xcolor} 

\usepackage[commentmarkup=uwave]{changes} 
\definechangesauthor[name={kaiwen}, color=Maroon]{kaiwen}
\definechangesauthor[name={xiaowei}, color=blue]{xiaowei}
\definechangesauthor[name={CHRIS}, color=magenta]{CHRIS}

\makeatletter
\let\NAT@parse\undefined
\makeatother
\usepackage[colorlinks, linkcolor=red, anchorcolor=red, citecolor=red]{hyperref}
\newcommand{\sysname}{{STUN}}   

\begin{document}
\maketitle
\thispagestyle{empty}
\pagestyle{empty}

\input{sections/0_abstract.tex}
\input{sections/1_introduction.tex}

\input{sections/2_related_work.tex}

\input{sections/3_method.tex}
\input{sections/4_experiment.tex}

\input{sections/5_results.tex}
\input{sections/6_conclusion.tex}

\bibliographystyle{IEEEtran}
\bibliography{my_library}

\end{document}

%% file: sections/0_abstract.tex
\begin{abstract}

    Place recognition is key to Simultaneous Localization and Mapping (SLAM) and spatial perception. However, a place recognition in the wild often suffers from erroneous predictions due to image variations, e.g., changing viewpoints and street appearance. 
    Integrating uncertainty estimation into the life cycle of place recognition is a promising method to mitigate the impact of variations on place recognition performance. However, existing uncertainty estimation approaches in this vein are either computationally inefficient (e.g., Monte Carlo dropout) or at the cost of dropped accuracy.
    This paper proposes \sysname, a self-teaching framework that learns to simultaneously predict the place and estimate the prediction uncertainty given an input image.
    To this end, we first train a teacher net using a standard metric learning pipeline to produce embedding priors. Then, supervised by the pretrained teacher net, a student net with an additional variance branch is trained to finetune the embedding priors and estimate the uncertainty sample by sample. During the online inference phase, we only use the student net to generate a place prediction in conjunction with the uncertainty. When compared with place recognition systems that are ignorant of the uncertainty, our framework features the uncertainty estimation for free without sacrificing any prediction accuracy.
    Our experimental results on the large-scale  Pittsburgh30k dataset demonstrate that \sysname\ outperforms the state-of-the-art methods in both recognition accuracy and the quality of uncertainty estimation. 

\end{abstract}

%% file: sections/1_introduction.tex
\section{Introduction}
Place recognition is an essential component of many robotics applications and enables downstream tasks such as metric localization and loop closure detection. 
While much effort has been made to improve recognition accuracy, few have considered the prediction uncertainty.

Uncertainty matters far more to place recognition than other components in robotics applications. In a SLAM application, for example, place recognition serves to correct the global pose graph, and 
a single false positive prediction 
may drastically 
lower the quality of
the estimated map and poses \cite{mur2017orb}. For this reason, it is desirable to know if a prediction can be trusted  -- a false positive prediction tends to have high uncertainty and should not be considered. Uncertainty is also inherent to real-world autonomous systems as it is inevitable for a place recognition system to produce uncertain results in the presence of changing illumination, viewpoints and object appearance. Uncertainty estimation is thus equally, if not more, important than an accurate prediction. 


\begin{figure}
    \centering
    \includegraphics[width=3.3in]{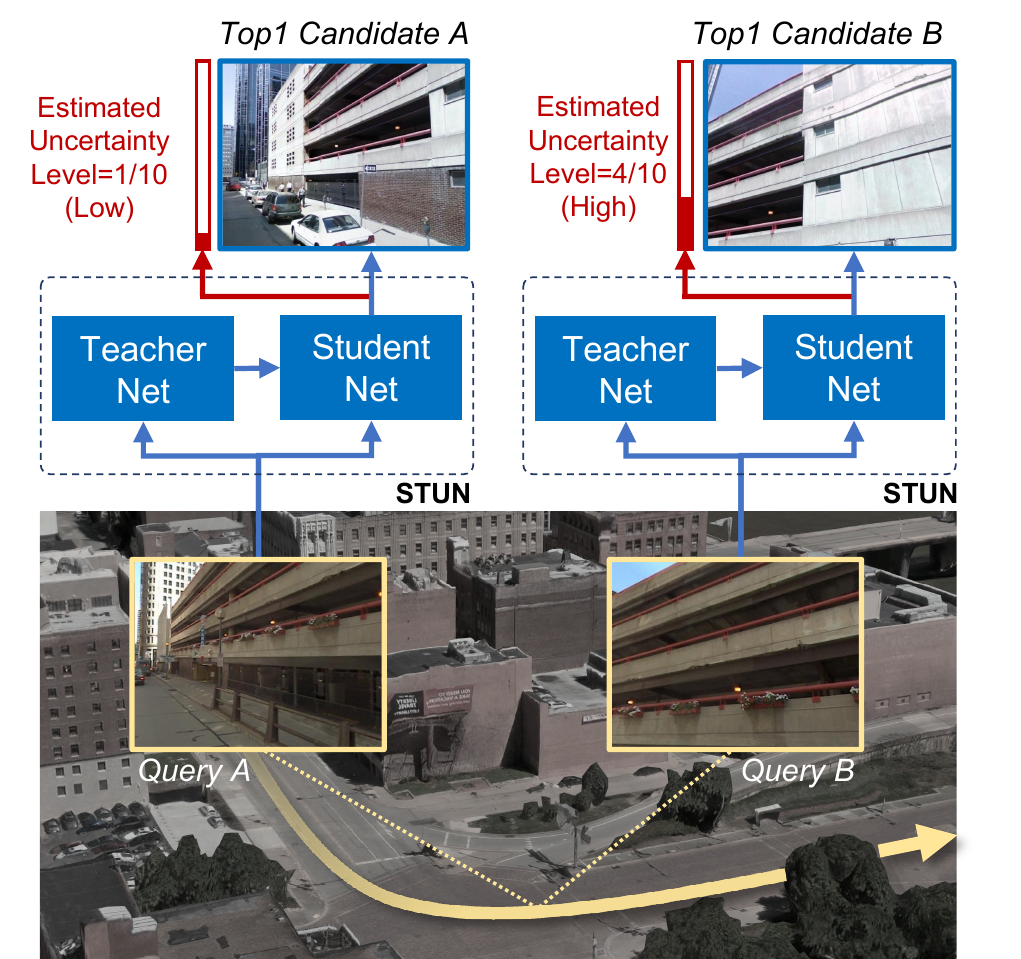}
    \caption{Traditional place recognition pipeline produces deterministic prediction. In contrast, Our \sysname\ uses a probabilistic prediction model, retriving the candidates (\textcolor{blue}{blue} arrow) while estimating the uncertainty of this retrival (\textcolor{red}{red} arrow), i.e., the probability of correctness. We found that samples with larger scene views (e.g., Query A has a larger scene view than Query B) tend to be estimated lower-uncertainty by \sysname.}
    \label{first}
\vspace{-0.7cm}
\end{figure}


Kendall et al. \cite{kendall2017uncertainties} address the epistemic and aleatoric uncertainty in deep learning and associate them with model and data, respectively. While no prior work 
estimates uncertainty for place recognition, a few efforts have been made to quantify the aleatoric or epistemic uncertainty for a similar task, image retrieval. Existing methods cast image retrieval as either classification\cite{chang2020data} or metric learning \cite{oh2018modeling, warburg2021bayesian} and then estimate uncertainty. Classification-based image retrieval is not of practical use since the number of classes increases linearly with the database's geographical coverage. For metric learning-based image retrieval methods, MC Dropout is used in \cite{oh2018modeling, loquercio2020general} to estimate epistemic uncertainty, which requires changes to the original network architecture and induces an expensive computational cost. 
To deal with this, BTL\cite{warburg2021bayesian} proposes a triplet likelihood to estimate aleatoric uncertainty. However, as shown in our experiments, its one-dimensional variance falls short for high-dimension embeddings in place recognition tasks and jeopardizes the recognition performance due to the 
change of loss functions. 
More importantly, 
the above methods are designed for image retrieval, which encodes images based on the similarity of their contents. The success cannot be directly transferred to place recognition, which concerns the geographical positions; hence, a positive pair in pairwise or triplet loss could present significant content variations. 
The above observations suggest that uncertainty estimation for place recognition is challenging. This paper focuses on the following questions:
\begin{itemize}
    \item[Q1] Can we estimate uncertainty for place recognition for free, i.e., \emph{without} compromising recognition accuracy?
    \item[Q2] Can an uncertainty estimation pipeline be 
    generic to all predominant metric learning loss functions? 
\end{itemize}

To answer the above two questions, 
we cast the uncertainty estimation for place recognition as a Knowledge Distilling (KD) task and particularly focus on estimating the aleatoric uncertainty. 
Specifically, a novel \textbf{s}elf-\textbf{t}eaching \textbf{un}certainty estimation pipeline dubbed \sysname\ (shown in Fig. \ref{first}) is proposed for metric learning-based place recognition: we first train a teacher net using a standard metric learning pipeline to generate embedding priors. Then a student net is trained to finetune the embeddings and simultaneously estimate the uncertainty for every instance. Only the student net is needed for the uncertainty quantification during the inference phase. 
Such a pipeline does not compromise the recognition accuracy (i.e., attempt to answer Q1) because we keep the student net as same as the original recognition network (teacher net), with the additional information from the 
teacher net being used only when computing the loss. Additionally, the teacher-student framework is intrinsically agnostic to loss functions (i.e., attempt to answer Q2) because only the student net is utilised for estimating uncertainty while the teacher net can be flexibly trained with different loss functions.
Experimental results show that by distilling the knowledge of the teacher net, the student net estimates a well-calibrated uncertainty and in the meantime has an improved 
recognition performance. 
The following are our contributions:
\begin{enumerate}
    \item \sysname\ is the first work on large scale uncertainty-aware place recognition by leveraging the Knowledge Distilling technique.
    \item \sysname\ achieves a better-calibrated uncertainty as well as a higher recognition accuracy compared with the existing methods for similar tasks. 
    \item \sysname\  is able to accommodate various loss functions predominantly used in place recognition, including pairwise, triplet, and quadruplet losses.
    \item The code of \sysname\ will be released to the community upon acceptance at \url{https://github.com/ramdrop/STUN}.
\end{enumerate}

%% file: sections/2_related_work.tex
\section{Related work}
\subsection{Place Recognition}
Localization, i.e., \emph{knowing where I am}, is always an essential ability for any autonomous moving robot. This capability is usually achieved by using GPS or SLAM. However, they are either unreliable or suffering from inevitable drift in the long run. Place recognition emerges as a cure, eliminating the drift in a large-scale environment for downstream localization tasks by recognizing a revisited place. 

Place recognition can be realized by different sensors, such as cameras\cite{arandjelovic2016netvlad}, LiDAR\cite{komorowski2021minkloc3d} or radar\cite{suaftescu2020kidnapped}. This paper will focus on visual place recognition since it shows the most prevalent usage. 
Early work on visual place recognition dates back to SeqSLAM\cite{milford2012seqslam}, where the author used handcrafted features to find the best-matched sequence from a geo-tagged database. Soon after  Convolutional Neural Network (CNN) showed compelling performance\cite{sunderhauf2015place}, a series of works on learning-based methods \cite{arandjelovic2016netvlad, chen2017only, zhu2018attention, oertel2020augmenting, hausler2021patch, garg2021seqnet} have been the dominating focus in place recognition.

\subsection{Uncertainty Estimation}
While already proved effective in making sensible predictions in various tasks, Deep Neural Network (DNN) will function more like a human brain if it provides confidence together with predictions. To quantify uncertainty in deep learning, Kendall et al. \cite{kendall2017uncertainties} 
associate 
epistemic uncertainty and aleatoric uncertainty with model and data, respectively. 
Based on Kendall's framework\cite{kendall2017uncertainties},
\cite{feng2018towards} 
quantifies uncertainty for LiDAR point cloud vehicle detection, and \cite{wang2018end} learns aleatoric uncertainty for visual odometry. Another line of estimating aleatoric uncertainty is based on variance propagation \cite{loquercio2020general, postels2019sampling}, but this framework requires an initial estimate of inherent sensor noise. For KD methodology, \cite{korattikara2015bayesian} applies KD to extract epistemic uncertainty from an ensemble of neural networks. In this paper, we use KD to estimate aleatoric uncertainty. 

\subsection{Uncertainty Estimation for Place Recognition}

Place recognition can induce unreliable predictions when confronted with degraded sensor data, e.g., blurred, overexposed or images with large viewpoint variations. 
Although predicting aleatoric uncertainty in classification or regression tasks\cite{kendall2016modelling} is straightforward, it is nontrivial to estimate it in place recognition 
because the embedding space is not well-defined by ground truth labels. 

Since there is no prior work on uncertainty estimation for place recognition, we summarise the research in its most relevant counterpart, image retrieval. Image retrieval is usually realized via classification or metric learning frameworks. On the one hand, Jie et al. \cite{chang2020data} frames the image retrieval as a classification task, converts it to a regression task, and then trains the network to learn the data uncertainty in these two tasks. But classification-based place recognition is not of practical use since the class number increases linearly with the database's geographical size.  On the other hand, PFE \cite{shi2019probabilistic} treats image retrieval as a metric learning task. The author trains a modified deterministic network to learn the data uncertainty and then uses the proposed Mutual Likelihood Score (MLS) to measure the probability of two images belonging to the same class. But it is limited to pairwise likelihood between positive samples and omits dissimilar pairs, which cannot assure optimum performance.
Taha et al. \cite{taha2019exploring,taha2019unsupervised} casts triplet loss as trivariate regression loss and then uses the framework of \cite{kendall2017uncertainties} to either learn aleatoric uncertainty or estimate model uncertainty using MC Dropout. 
However, the modified triplet loss function might lead to sub-optimal performance.
BTL outperforms \cite{taha2019exploring,taha2019unsupervised} by deriving a likelihood for triplet loss, which is then used to train the network. But the proposed complex likelihood loss function requires more attention to assure the training convergence and cannot consistently achieve comparable recognition performance as a standard triplet metric learning method. Our work is inspired by \cite{poggi2020uncertainty}, which uses a teacher-student framework to learn pixel depth uncertainty. Though proved effective in self-supervised depth estimation, it remains unknown how it can be utilized to solve the challenging estimate uncertainty problem in place recognition, which features totally different inputs, neural network architectures, and goals.


%% file: sections/3_method.tex
\section{Method}
\subsection{Preliminary}
\label{section_preliminary}

By retrieving the best-matched candidate from a geo-tagged database when given a query sample, a place recognition system predicts a coarse location of the query. The methods to achieve this goal can be classified into two categories: classification-based and metric learning based. A classification-based place recognition pipeline\cite{chen2017deep} geographically divides the database into a certain number of classes and then predicts which class a query belongs to. But to maintain a consistent localization precision, the minimum number of classes will increase linearly with the database's geographical area. This limitation makes classification-based place recognition impractical for real-world large-scale applications. Thus, in this study, we only focus on metric learning-based place recognition.

In a metric learning based place recognition setting\cite{uy2018pointnetvlad,arandjelovic2016netvlad}, one will first map samples from image space $\mathcal{X}$ to embedding space $\mathcal{Y}$, and then uses nearest neighbor search to conduct a retrieval. Formally, images are represented as $\{ \mathbf{x}_i \in \mathbb{R} ^{C \times H \times W}\mid i=1,2,...,N \}$ and embeddings as $\{{\mathbf{y}_i} \in \mathbb{R} ^D \mid i=1,2,...,N\}$. Every embedding is a deterministic vector that can be regarded as a high-dimensional point in the embedding space. The mapping function $f_{\mu}: \mathcal{X} \to \mathcal{Y} $ is represented by a DNN and formulated as:
\begin{equation}
\label{mapping}
{\mathbf{y}_i} = f_{\mu}({\mathbf{x}_i})
\end{equation}
Since there are no ground truth labels for embeddings, the DNN is trained by minimizing the distance between a query (anchor) embedding and its positive embedding while maximizing the distance between a query embedding and its negative embedding. During the inference phase, the nearest neighbor search in embedding space is used to retrieve the best-matched candidate to the query.

\subsection{Probabilistic Place Recognition Pipeline}



Due to the variations in the real world when collecting images, including changing viewpoints, exposure and street appearance, neither images nor embeddings are promised always well to represent a place. For example, an image of a landmark would be more characteristic than that of a plain sidewalk. However, a conventional place recognition pipeline will always cluster embeddings of images from the same place, even if some of them contribute little information to a place. This ignorance of image variations corrupts the embedding space. 

To capture such variations, we propose to introduce \emph{probabilistic embedding model}, which describes an embedding as a distribution rather than a point. As Gaussian model shows good performance in metric learning \cite{warburg2021bayesian}, we 
here adopt a Gaussian distribution, i.e.,
\begin{equation}
    \mathbf{y_i^{\prime}}\sim \mathcal{N} \left( f_{\mu}({\mathbf{x_i}}), f_{\sigma} (\mathbf{x_i})\right)
\end{equation}
where $f_{\sigma}(\cdot)$ represents a learnable variance prediction network. 
The motivation behind the 
probabilistic embeddings is that the learned variance acts like a predictor telling how well an embedding represents a place. For those images that contribute little to a place's character, our probabilistic place recognition pipeline will not brutally push their embeddings close to other embeddings of the same place. In the Sec.\ref{section_results} we show that probabilistic embedding model helps learn a better mapping function $f_{\mu}$.

\subsection{Self-Teaching Uncertainty Estimation (\sysname)}
Existing methods\cite{taha2019exploring, warburg2021bayesian} learn probabilistic embeddings for metric learning, but their recognition accuracy is degraded compared to their deterministic counterparts due to their modifications of either network architectures or loss formulations. To avoid performance degradation,
we utilize KD and only extract uncertainty from embeddings.
In this way, \emph{no modification to network or loss} is required, and the original recognition performance is preserved (or improved).
Specifically, we adopt a teacher-student network and first train a teacher net to output embedding mean priors. Then, under the teacher net's supervision, we train a student net to finetune the embedding mean and in the meantime estimate the embedding variance.
\subsubsection{Teacher Net}
\label{sec_teacher_net}

We denote the feature extractor of the teacher net as $f_{T,o}(\cdot)$, and the mean head as  $f_{T,\mu}(\cdot)$. Given an input image $\mathbf{x}_i$, the teacher net predicts the embedding mean $\boldsymbol{{\mu}}_{T,\mathbf{x}_i}$ by using:
\begin{equation}
    \boldsymbol{{\mu}}_{T,\mathbf{x}_i} = f_{T,\mu}(f_{T,o}(\mathbf{x}_i))
\end{equation}
For easy comparison with the state-of-the-arts (SOTA), our feature extractor backbone $f_{T,o}(\cdot)$ follows \cite{warburg2021bayesian}, consisting of layers of ResNet50 before the global average pooling layer and a GeM pooling layer\cite{radenovic2018fine}. The mean head $f_{T,\mu}(\cdot)$ is a L2 normalization layer, $\mathbf{x}_i \in \mathbb{R} ^{3 \times 200 \times 200} $ and $\boldsymbol{{\mu}}_{T,\mathbf{x}_i} \in \mathbb{R} ^{2048}$.


Note that unlike existing methods that are limited to pairwise\cite{shi_probabilistic_2019} or triplet loss\cite{warburg2021bayesian}, our teacher net is a generic uncertainty estimation framework that can be trained using any metric learning loss, such as contrastive loss\cite{hadsell2006dimensionality}, triplet loss\cite{schroff2015facenet}, quadruplet loss\cite{chen2017beyond}, 
angular loss\cite{wang2017deep} and margin loss\cite{wu2017sampling}. 
This study will show the flexibility of \sysname\ by training our teacher net using the most common losses, including contrastive, triplet, and quadruplet losses. We briefly summarise the principles of these losses for completeness.

\textbf{Contrastive Loss} was first proposed by Raia et al. \cite{hadsell2006dimensionality} to learn a discriminative mapping that can distinguish between similar and dissimilar pairs. Given an input doublet images $\{(\mathbf{x}_i, \mathbf{x}_j)\}$, the loss function is formulated as follows:
\begin{flalign}
    \label{eq_cont}
    & \mathcal{L}_{cont} = \mathds{1}_{\{\mathbf{x}_i, \mathbf{x}_j\}}  \cdot d^2(\boldsymbol{\mu}_{T,\mathbf{x}_i}, \boldsymbol{\mu}_{T,\mathbf{x}_j}) \nonumber \\
    & \quad \quad \quad + \overline{\mathds{1}}_{\{\mathbf{x}_i, \mathbf{x}_j\}} \cdot (m-d^2(\boldsymbol{\mu}_{T,\mathbf{x}_i}, \boldsymbol{\mu}_{T, \mathbf{x}_j}))
\end{flalign}
where $\mathds{1}\{ \cdot \}$ is an indicator function such that it is $1$ when $\mathbf{x}_i$ and $\mathbf{x}_j$ are a ground truth similar pair, $0$ otherwise. $d$ means Euclidean distance, and $m$ a predefined constant margin.

\textbf{Triplet Loss}\cite{schroff2015facenet} could be more efficient in learning a discriminative embeddings as it involves three samples at a time, while contrastive loss takes two samples as input. The triplet loss is given as follows:
\begin{equation}
    \label{triplet_loss}
    \mathcal{L}_{tri}= \max\{d(\boldsymbol{\mu}_{T, \mathbf{x}_a}, \boldsymbol{\mu}_{T, \mathbf{x}_p}) - d(\boldsymbol{\mu}_{T, \mathbf{x}_a}, \boldsymbol{\mu}_{T, \mathbf{x}_n}) + m, 0\}
\end{equation}
where  $\mathbf{x}_a, \mathbf{x}_p, \mathbf{x}_n$ are the anchor, positive and negative sample, respectively.

\textbf{Quadruplet Loss}\cite{chen2017beyond} aims to futher enlarge inter-class distance for metric learning by introducing an additional negative sample on top of the triplet loss. The quadruplet loss is given as follows:
\begin{flalign}
    \label{eq_quad}
    & \mathcal{L}_{quad}= \max\{d(\boldsymbol{\mu}_{T, \mathbf{x}_a}, \boldsymbol{\mu}_{T, \mathbf{x}_p}) - d(\boldsymbol{\mu}_{T, \mathbf{x}_a}, \boldsymbol{\mu}_{T, \mathbf{x}_{n1}}) + m_1, 0\} \nonumber  \\
    & \quad \quad + \max\{d(\boldsymbol{\mu}_{T, \mathbf{x}_a}, \boldsymbol{\mu}_{T, \mathbf{x}_p}) - d(\boldsymbol{\mu}_{T, \mathbf{x}_a}, \boldsymbol{\mu}_{T, \mathbf{x}_{n2}}) + m_2, 0\}
\end{flalign}
where $\mathbf{x}_a, \mathbf{x}_p$ are the anchor and positive sample, respectively, $\mathbf{x}_{n1}$ is the first negative sample to the anchor, and $\mathbf{x}_{n2}$ is the second negative sample dissimilar to any other samples in the quadruplet.  $m_1$ and $m_2$ are predefined constant margins.



\subsubsection{Student Net}
Once trained, the teacher net will produce a deterministic embedding for each sample in the dataset. But the deterministic training of the teacher net does not consider the inherent aleatoric uncertainty, rendering the teacher net a sub-optimal mapping function. In this regard, we take the mapping function represented by the teacher net as a prior and train a student net to finetune the prior while learning an aleatoric uncertainty simultaneously.

\begin{figure}
    \centering
    \includegraphics[width=2.6in]{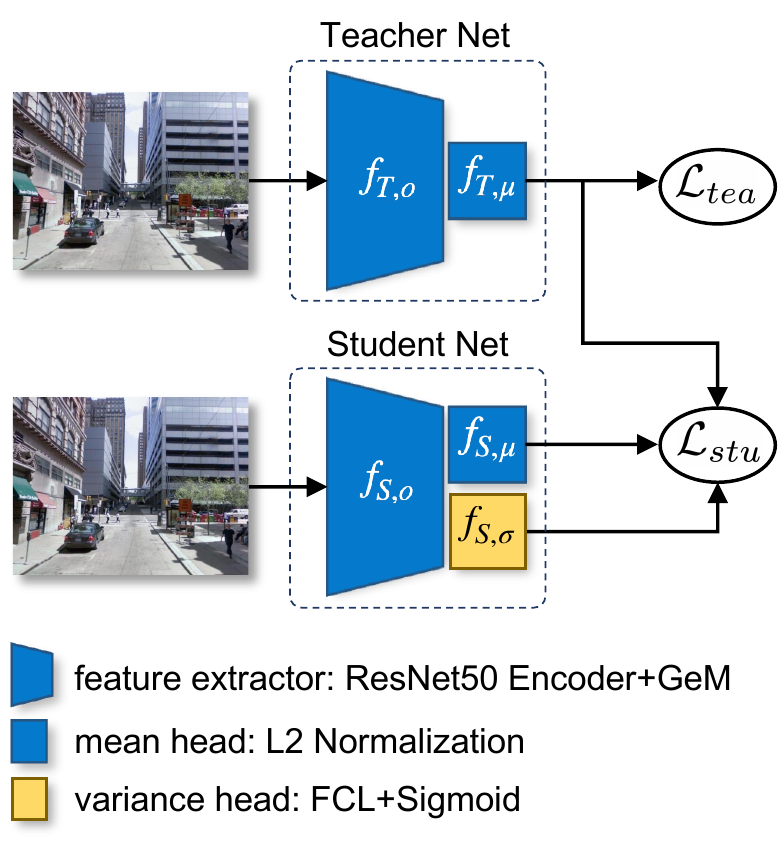}
    \caption{The proposed \sysname: we first train a teacher net using $\mathcal{L}_{tea}$, which can take the form of various metric learning loss, e.g., $\mathcal{L}_{cont}$, $\mathcal{L}_{tri}$ and $\mathcal{L}_{quad}$ (see Sec.\ref{sec_teacher_net} for details). Then, under the supervision of the teacher net, the student net with an additional variance head is trained to finetune the embeddings and estimate the uncertainty simultaneously. }
    \label{pipeline}
\vspace{-0.5cm}    
\end{figure}

The student net shares the same feature extractor as the teacher net, except that, to capture the aleatoric uncertainty, an additional variance head is added in parallel to the mean head. Specifically, we denote the feature extractor of the student net as $f_{S,o}(\cdot) = f_{T,o}(\cdot)$, the mean head as  $f_{S,\mu}(\cdot)=f_{T,\mu}(\cdot)$, and the variance head as $f_{S,\sigma}(\cdot)$. The variance head $f_{S,\sigma}(\cdot)$ comprises a fully connected layer and a sigmoid activation layer.
Given an input image $\mathbf{x}_i$, the embedding $\boldsymbol{{\mu}}_{S,\mathbf{x}_i}$ is predicted by the student net using:
\begin{equation}
    \boldsymbol{{\mu}}_{S,\mathbf{x}_i} = f_{{S},\mu}(f_{{S},o}(\mathbf{x}_i))
\end{equation}
\begin{equation}
    \boldsymbol{{\sigma}}_{S,\mathbf{x}_i} = f_{{S},\sigma}(f_{{S},o}(\mathbf{x}_i))
\end{equation}
where $\boldsymbol{{\mu}}_{S,\mathbf{x}_i}, \boldsymbol{{\sigma}}_{S,\mathbf{x}_i} \in \mathbb{R} ^{2048}$.


The student net is trained using the  proposed uncertainty-aware loss:
\begin{equation}
    \label{stu_loss}
    \mathcal{L}_{stu}=\sum _i^N \frac{(\boldsymbol{\mu}_{S, {\mathbf{x}_i}}-{\boldsymbol{\mu}}_{T,\mathbf{x}_i})^2}{2\boldsymbol{\sigma}^2_{S, \mathbf{x}_i}}+\frac{1}{2}\ln\boldsymbol{\sigma}^2_{S, \mathbf{x}_i}
\end{equation}
Recall that ${\boldsymbol{\mu}}_{T,\mathbf{x}_i}$ is the embedding prior provided by the teacher net. During training, the numerator of the first term will push the learned embedding close to the embedding prior. For the embedding that is far from its prior, a large variance $\boldsymbol{\sigma}^2_{S,\mathbf{x}_i}$ will be assigned to the denominator, avoiding the network being updated by an unreasonable large gradient. The last term of  $\mathcal{L}_{stu}$ functions as a regulariser, preventing the network from always outputting a large variance.

A teacher-student framework is usually used to distill knowledge from a complex network, but in contrast, we use it as a fine-tuning and uncertainty estimation strategy. Since the teacher net and the student net share the same feature extractor architecture, and no additional ground truth information is needed to enforce the uncertainty-aware training, we refer to our teacher-student framework as a \emph{self-teaching} framework. The overall self-teaching uncertainty estimation pipeline is shown in Fig. \ref{pipeline}.

\textbf{Comparison with BTL\cite{warburg2021bayesian} and PFE\cite{shi_probabilistic_2019}}. We compare \sysname \ with two SOTA techniques, BTL\cite{warburg2021bayesian} and PFE\cite{shi_probabilistic_2019}. BTL \cite{warburg2021bayesian} approximates the likelihood of a standard triplet loss as an isotropic Gaussian distribution, but its one-dimensional variance could fall short in capturing uncertainties in place recognition high-dimension embedding space (shown in Sec.\ref{section_results}). Besides, its customized loss function cannot assure a consistent performance as a standard triplet loss \cite{warburg2021bayesian}. PFE\cite{shi_probabilistic_2019} builds upon a pretrained embedding mean predictor and then learns aleatoric uncertainty by maximizing the likelihood of two similar embeddings being a positive pair. However, omitting dissimilar embeddings would lead to a sub-optimum uncertainty estimation. In contrast, the teacher net of \sysname\ is trained using a standard triplet loss, this empowers the student net to achieve an on-par recognition performance as a standard triplet loss. When learning aleatoric uncertainty, the student net is trained over all samples, making it learn an optimum portrayal of aleatoric uncertainty. In addition, the teacher net of \sysname\ can be adapted to any metric learning loss dependent on ad-hoc tasks, rendering it a more flexible solution than the existing methods for general real-world applications. 


%% file: sections/4_experiment.tex
\section{Experiments Setup}
\subsection{Dataset}
We evaluate \sysname\ and SOTA methods on \textbf{Pittsburgh}\cite{torii2013visual} dataset, which is a large image database from Google Street View. The original dataset contains 250k training samples and 24k test query samples. To facilitate training, we follow the split of NetVLAD\cite{arandjelovic2016netvlad}, using a smaller subset that consists of 10k samples in the training/validation/test split. All the images are taken at different times in the day, years apart.


\subsection{Implementation details}
After searching hyper-parameters on the training and validation set, we set margin  $m=0.4$ for contrast loss,  $m=0.1$ for triplet loss, and  $m_1=0.1, m_2=0.1$ for quadruplet loss. For the network training, we use a batch size of $8$, Adam optimizer with an initial learning rate $1\times 10^{-5}$ which is decayed by $0.99$ every epoch, and weight decay  $0.001$. We follow \cite{musgrave2020metric} and freeze BatchNorm parameters during training to reduce overfitting. Following the scale of NetVLAD\cite{arandjelovic2016netvlad}, we regard places in the database that are within the radius=$\SI{10}{\m}$ to the query as true positives, while those outside the radius=$\SI{25}{\m}$ area as true negatives. We 
follow \cite{warburg2021bayesian} and adopt a hard negative mining strategy when training the teacher net, where we only input the model with doublets, triplets and quadruplets that violate its loss constraints Eq. \ref{eq_cont}, Eq. \ref{triplet_loss}, and Eq. \ref{eq_quad}.

\subsection{Evaluation Metrics}
We evaluate the performance of uncertainty-aware place recognition from two aspects: recognition performance and uncertainty estimation performance.

\textbf{Recognition performance} is evaluated by r@N\cite{arandjelovic2016netvlad} (recall@N), mAP@N\cite{musgrave2020metric} (mean Average Precision) and AP\cite{hou2018evaluation} (Average Precision). Recall@N denotes the percentage of true positives to the retrieved top $N$ candidates. mAP@N measures the precision of the retrieved top $N$ candidates. AP can be regarded as the area under precision-recall curve.

\textbf{Uncertainty estimation performance} is evaluated by ECE (expected calibration error) \cite{warburg2021bayesian}. ECE illustrates the correlation between uncertainty level and recognition performance. Ideally, an uncertainty-aware place recognition system generates the best recognition performance 
for the lowest uncertainty samples and shows degraded recognition performance as the uncertainty grows. To calculate ECE for r@N, we first divide the recognition results of all test queries into $M$ equal bins $\{B_i|i=1,2,.., M\}$ based on uncertainties\footnote{Similar to \cite{chang2020data}, we use the mean of variance along all embedding dimensions as an approximation of uncertainty level.  
} of the query samples, and then calculate the r@N for queries in each bin. $M$ bins will result in $M$ uncertainty levels $\mathcal{U}(B_i)$, which are normalized such that the maximum uncertainty level equals $1.0$. In order to calculate ECE, a confidence level $\mathcal{C}$ is defined as $\mathcal{C} = 1.0 - \mathcal{U}$. The final ECE for r@N score is obtained by
\begin{equation}
ECE_{r@N} =  \frac{\sum _i^M \vert B_i \vert \cdot \vert r@N(B_i) - \mathcal{C}(B_i)\vert}{\sum _i^M \vert B_i \vert}
\end{equation}
Intuitively, the term $\vert r@N(B_i) - \mathcal{C}(B_i)\vert$ encourages recognition accuracy to match confidence level. Replacing r@N with mAP@N will generate $ECE_{mAP@N}$ and $ECE_{AP}$. We use reliability diagram\cite{guo2017calibration} as a qualitative metric. 

%% file: sections/5_results.tex
\section{Results}
\label{section_results}
\begin{figure}
  \centering
  \includegraphics[width=2.6in]{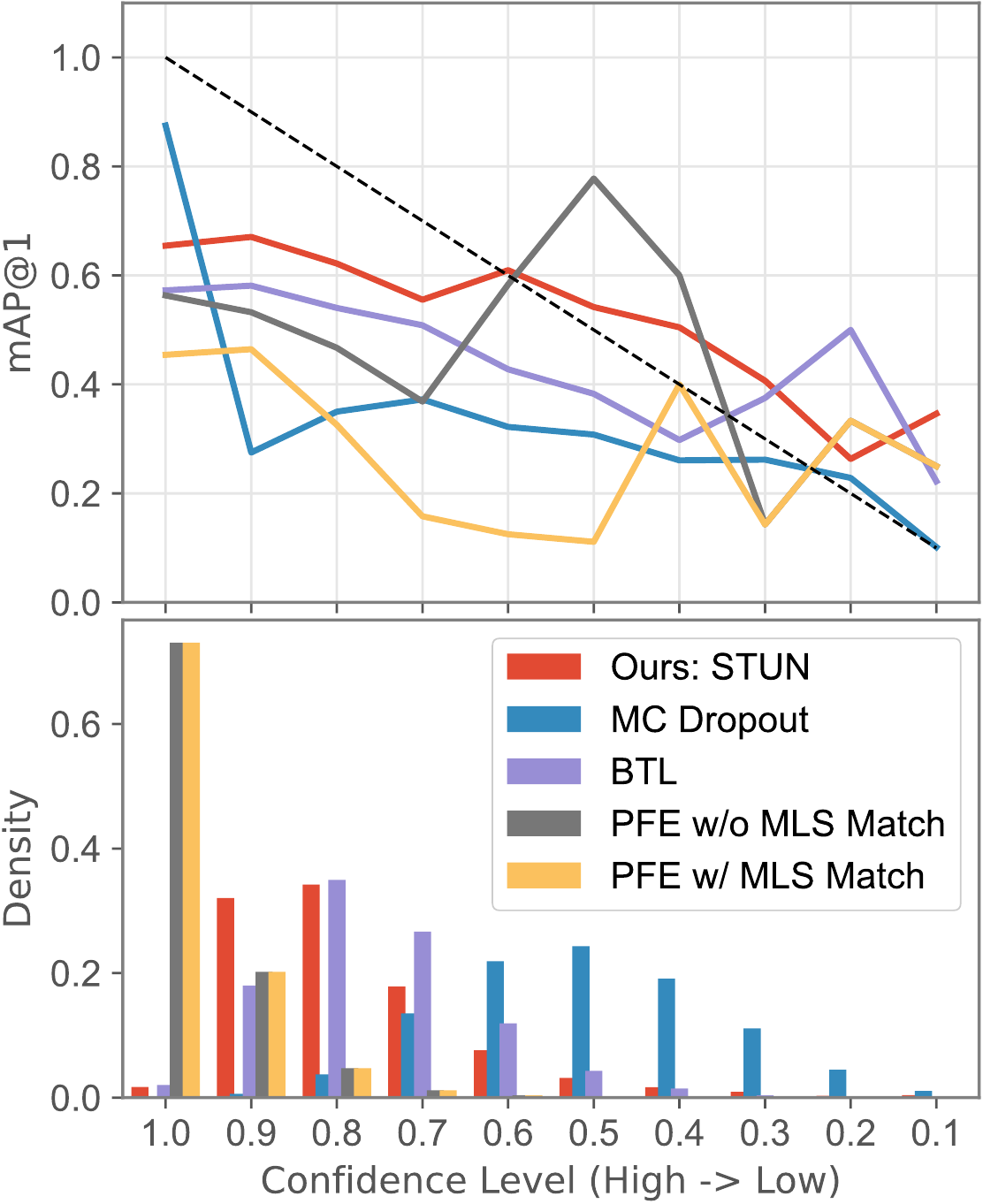}
  \caption{Reliability diagram for \sysname \ and SOTA methods. \sysname\  produces the best calibrated uncertainty.}
  \label{fig_ece}
\end{figure}

\begin{figure}
  \centering
  \includegraphics[width=2.6in]{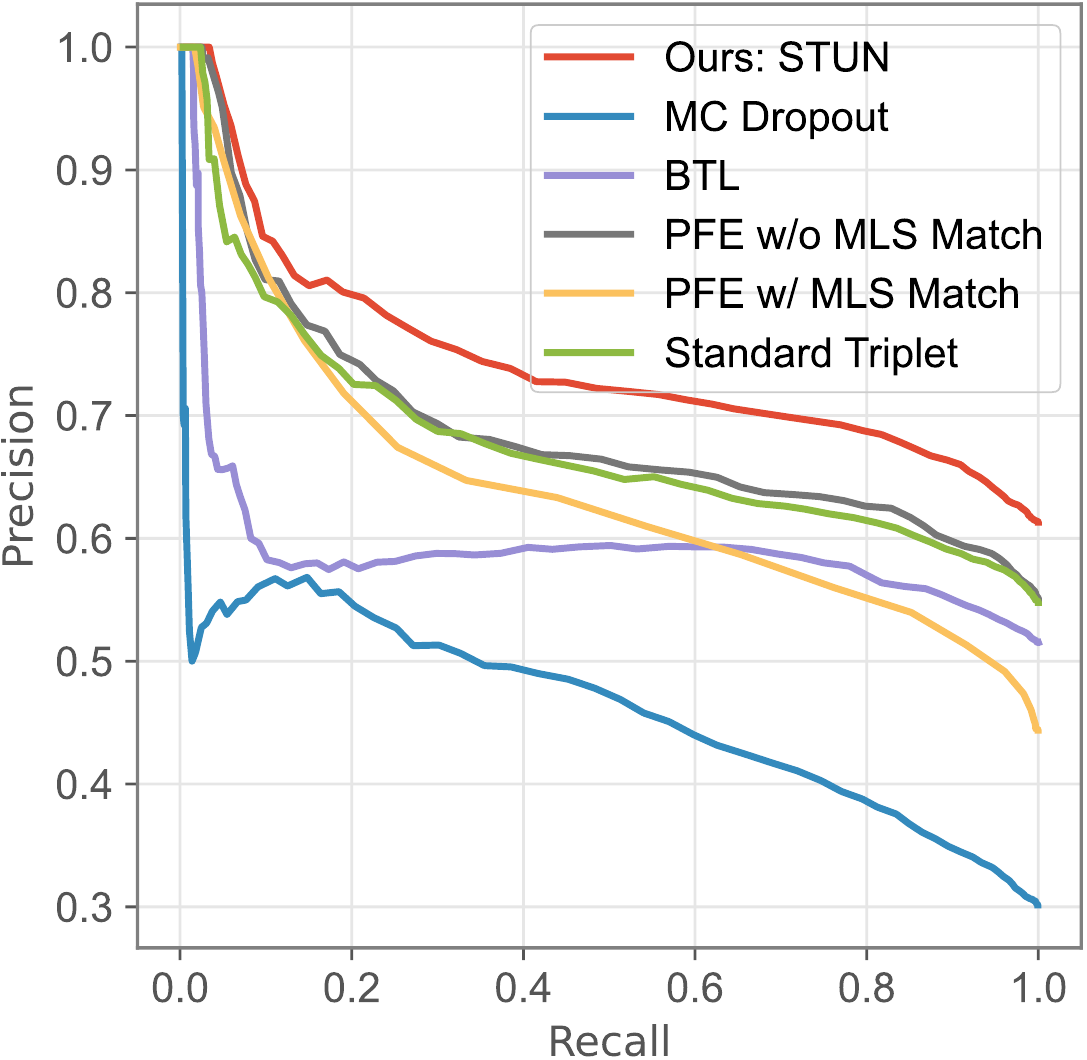}
  \caption{
    Recall and precision curve for \sysname \ and SOTA methods. \sysname\ shows the best recognition performance. 
  }
  \label{fig_pr}
\vspace{-0.6cm}  
\end{figure}

\begin{figure}
  \centering
  \includegraphics[width=2.6in]{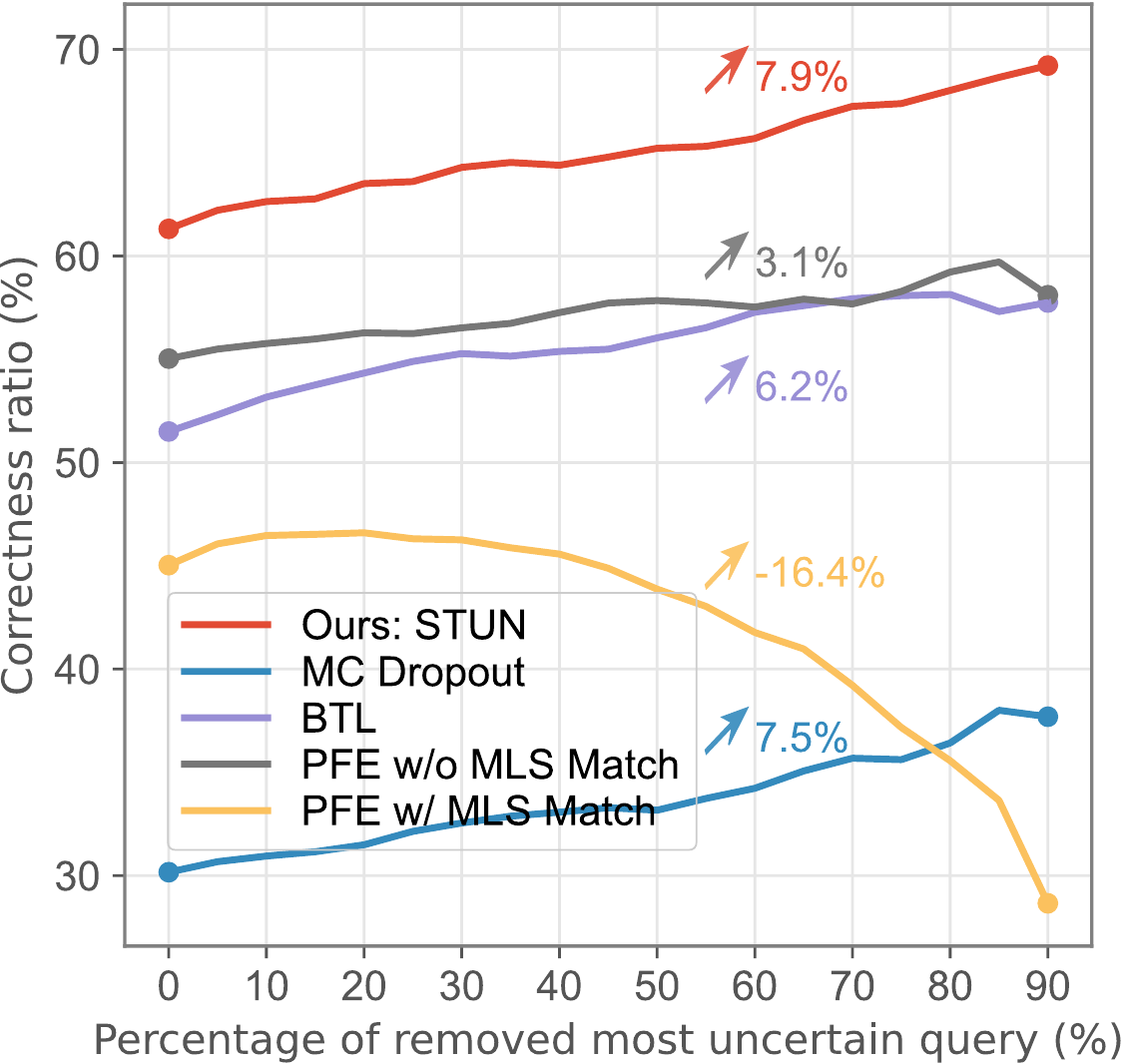}
  \caption{Recognition performance increases when most uncertain queries are removed for \sysname\ and SOTA methods. The numbers above the curves denote the improvement between the starting point and end point. \sysname\ shows the largest recognition improvement when the same amount of uncertain samples is discarded. 
  }
  \label{fig_cleaning_query}
\vspace{-0cm}  
\end{figure}

\begin{figure}
  \centering
  \includegraphics[width=3.3in]{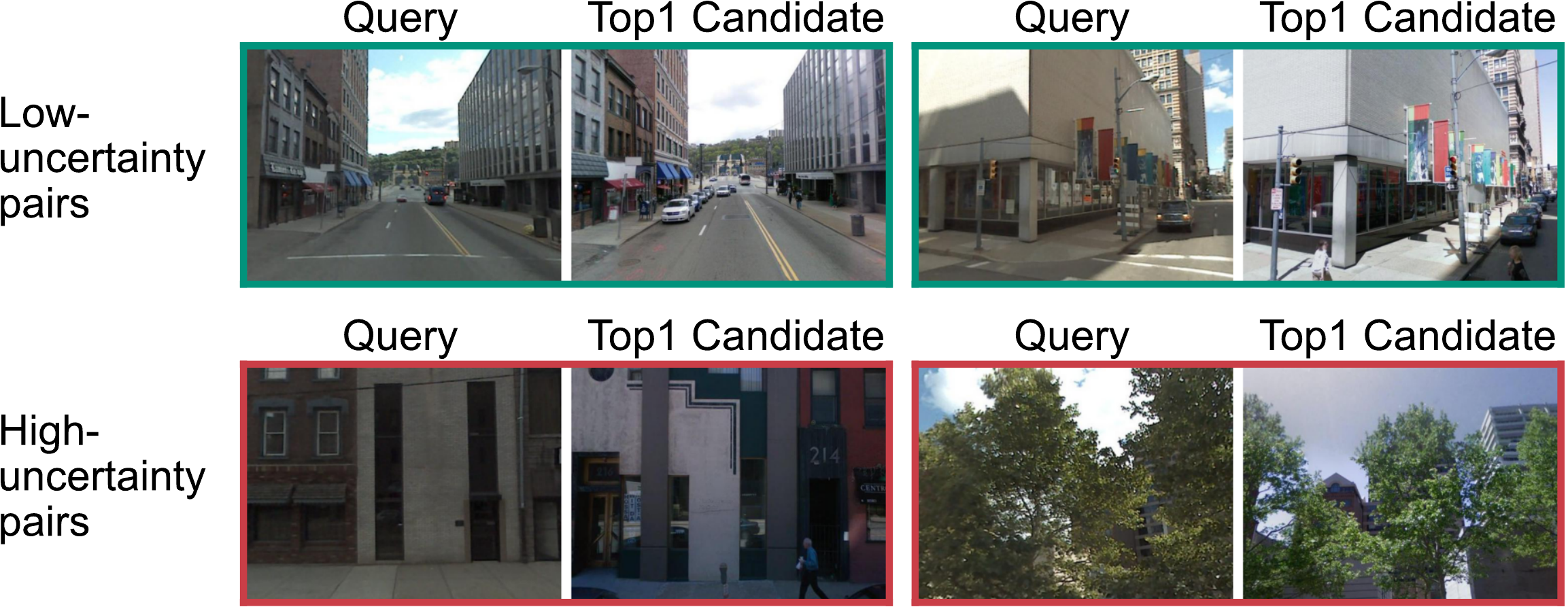}
  \caption{Pairs between query and its top-1 candidate which \sysname\ assesses as low-uncertainty pairs (top row) and high-uncertainty pairs (bottom row). \textcolor{RightColor}{Green} means the retrived top-1 candidate is a true positive, while \textcolor{WrongColor}{red} false positive. We observe low-uncertainty pairs tend to have larger scene views than high-uncertainty ones.
  }
  \label{fig_pair}
\vspace{-0.6cm}  
\end{figure}

\begin{table*}[!htb]
    \centering
    \caption{Recognition performance and uncertainty quality on the Pittsburgh dataset}
    \begin{threeparttable}

    \begin{tabular}{c|cccccc}
    \label{performance_table}
                      & \multicolumn{1}{c}{r@1/5/10 ↑}    & \multicolumn{1}{c}{mAP@1/5/10 ↑}  & \multicolumn{1}{c}{AP ↑} & \multicolumn{1}{c}{$ECE_{r@1/5/10}$}↓  & \multicolumn{1}{c}{$ECE_{mAP@1/5/10}$↓} & \multicolumn{1}{c}{$ECE_{AP}$↓}  \\ 
    \hline\hline
     Standard Triplet    & 0.548 / 0.802 / 0.877          & 0.548 / 0.268 / 0.167          & 0.678                  & -   & -              & -                        \\

    MC Dropout\cite{taha2019exploring}    & 0.302 / 0.523 / 0.611          & 0.302 / 0.108 / 0.061          & 0.463                  & 0.205 / 0.075 / 0.112   & 0.205 / \underline{\textbf{0.396}} / \underline{\textbf{0.443}}              & 0.111                        \\
    PFE w/o MLS \cite{shi_probabilistic_2019}      & 0.550 / 0.805 / 0.876          & 0.550 / 0.266 / 0.167          & 0.690                  & 0.413 / 0.160 / \underline{\textbf{0.092}}            & 0.413 / 0.695 / 0.795              & 0.275                        \\
    PFE w/ MLS \cite{shi_probabilistic_2019}      & 0.444 / 0.680 / 0.764          & 0.444 / 0.199 / 0.120          & 0.655                  & 0.517 / 0.284 / 0.200            & 0.517 / 0.762 / 0.841              & 0.336                        \\
    BTL\cite{warburg2021bayesian}      & 0.515 / 0.766 / 0.840          & 0.515 / 0.252 / 0.158          & 0.591                  & 0.236 / \underline{\textbf{0.058}} / 0.105   & 0.236 / 0.497 / 0.591              & 0.131                        \\ 
    Ours: \sysname\     & \textbf{0.613}\tnote{1} / \textbf{0.840} / \textbf{0.898}          & \textbf{0.613} / \textbf{0.280} / \textbf{0.171}          & \textbf{0.739}                  & \textbf{0.171} / 0.084 / 0.127   & \textbf{0.171} / 0.491 / 0.600     & \textbf{0.067}                        \\
    \midrule
    Ours: \sysname\ (Contrastive) & 0.512 / 0.767 / 0.845          & 0.512 / 0.204 / 0.119          & 0.610                & 0.185 / 0.086 / 0.152            & 0.185 / 0.493 / 0.577              & \underline{0.054}                        \\
    Ours: \sysname\ (Quadruplet)  & \underline{0.625}\tnote{2} / \underline{0.846} / \underline{0.902}  & \underline{0.625} / \underline{0.294} / \underline{0.180}  & \underline{0.740}    & \underline{0.114} / 0.140 / 0.194            & \underline{0.114} / 0.412 / 0.524              & 0.072                      
    \end{tabular}
    \begin{tablenotes}
        \footnotesize
        \item[1] \textbf{bold} denotes the best performance among methods trained with triplet inputs (methods in the top part).
        \item[2] \underline{underline} denotes the best performance among all methods.
      \end{tablenotes}    
\end{threeparttable}     

\end{table*}

\begin{figure*}
  \centering
  \includegraphics[width=6.5in]{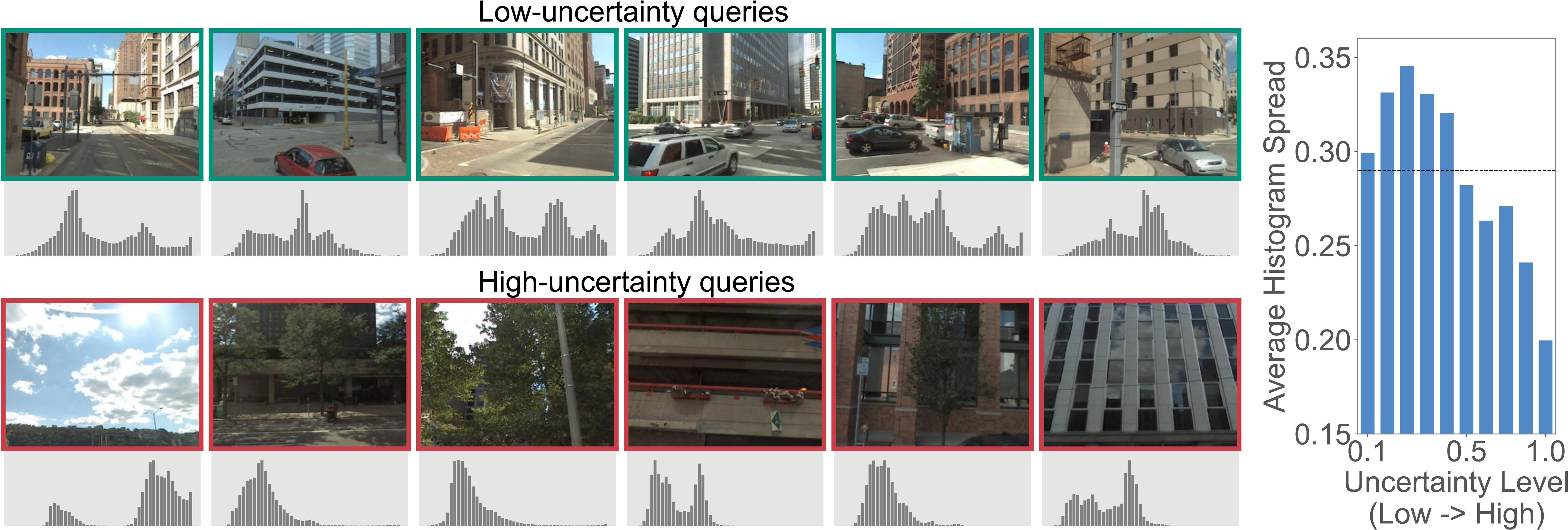}
  \caption{On the left are query samples which \sysname\ assesses as low-uncertainty (first row) and high-uncertainty (third row) ones. Histograms (Y-axis: number of pixels vs. X-axis: pixel intensity) of their greyscale images are shown below samples. \textcolor{RightColor}{Green} border means the retrieved closest candidate is a true positive, while \textcolor{WrongColor}{red} denotes false positive. It can be observed that query samples with large scene views tend to have low uncertainties, while those with limited scene views are likely to have high uncertainties. On the right is the relation between Histogram Spread (HS) \cite{tripathi2011performance} and query uncertainty, which shows that query images with lower uncertainties have higher HS. This trend can also be found in the histograms under each query image, where a high HS means a flat histogram (second row), while a low HS means a peaky histogram (last row).
  } 
  \label{fig_query}
\vspace{-0.4cm}  
\end{figure*}

We compare \sysname\ with the following SOTA methods for uncertainty-aware prediction:

\textbf{MC Dropout}\cite{taha2019exploring}: we follow the original paper's parameters by employing dropout layers after every convolutional layer in our ResNet50 backbone and setting dropout rate $p=0.2$ during both the training and inference phase. During the inference phase, forward propagation is performed 40 times as suggested by \cite{kendall2016modelling}. We note that MC Dropout is designed for epistemic uncertainty estimation rather than the aleatoric uncertainty targeted by STUN. The reason we still have it as our baseline is for a comprehensive comparison of place recognition across uncertainty estimation categories. 

\textbf{PFE}\cite{shi_probabilistic_2019}: PFE first trains a deterministic embedding mean head and then freezes it. After that, it only trains the variance head. To make a fair comparison, we only substitute our $\mathcal{L}_{stu}$ with MLS loss\cite{shi2019probabilistic} when training the variance head. Once the variance head is trained, we report the performance of PFE without MLS matching as \textbf{PFE w/o MLS}, and the one with MLS matching as \textbf{PFE w/ MLS}.

\textbf{BTL}\cite{warburg2021bayesian}: We follow the parameters of the original paper without extra modification for a fair comparison.

Since MC Dropout, PFE and \sysname\ use a high dimensional variance, we map it to a scalar by averaging variances over all dimensions.

\subsection{Quantitative Analysis}

Table. \ref{performance_table} shows recognition performance and uncertainty quality of different methods on the Pittsburgh dataset.

\textbf{Place Recognition Performance}:  It is clear that when trained with triplet inputs, \sysname\ achieves the highest recognition performance with r@1=$0.613$ among all. PFE w/o MLS achieves the second-highest performance, while PFE w/ MLS only obtains an r@1=$0.444$, suggesting that recognition performance could be degraded by inferior uncertainty estimation. This is followed by BTL with r@1=$0.515$. BTL uses a triplet likelihood to train the network, which however cannot assure a comparable recognition performance as a standard triplet loss\cite{schroff2015facenet}. MC Dropout has the worst recognition performance because the dropout on convolutional layers significantly decreases the network's representation capability. The same trend can also be observed from Fig. \ref{fig_pr}, where \sysname\ shows the best recognition performance. It is worth noting that \sysname\ even outperforms a standard triplet pipeline in recognition performance. This can be attributed to the fact that \sysname\ utilizes a standard triplet network as a prior and finetunes it with learned aleatoric uncertainty.

\textbf{Uncertainty Quality}: Fig. \ref{fig_ece} shows the reliability diagram for methods trained with triplet inputs. It is shown that the \sysname\ curve is closest to the ideally calibrated line (dotted line). Table. \ref{performance_table} presents the quantitative results of uncertainty estimation. With respect of $ECE_{r@1}$, $ECE_{mAP@1}$ and ${ECE_{AP}}$, \sysname\ produces the best-calibrated embedding uncertainty among methods trained with triplet inputs, while PFE w/ or w/o MLS shows the worst uncertainty quantification with the largest $ECE_{mAP@1}$ scores. In terms of $ECE_{r@5/10}$, \sysname\ obtains a small ECE value indicating the uncertainty is well-calibrated, which is marginally outperformed by the BTL. When comparing $ECE_{mAP@5/10}$, we found MC Dropout delivers the best uncertainty estimation quality while \sysname\ ranks second place. However, MC Dropout achieves this at the cost of drastically losing recognition performance r@1 by $0.246$. 
In summary, by leveraging a self-teaching strategy, \sysname\ not only estimates the best-calibrated uncertainty with $ECE_{mAP@1}=0.171$, but also achieves the highest recognition performance with r@1=$0.613$.

\textbf{Triplet vs. Contrastive vs. Quadruplet Loss}: We present the performance of \sysname\ trained with contrastive loss and quadruplet loss in Table. \ref{performance_table}. We write \sysname\ when training with the default triplet loss and write \sysname\ (Contrastive) and \sysname\ (Quadruplet) to refer to specific losses. 
We observe that 
\sysname\ (Contrastive) performs worse than \sysname\ in both recognition and uncertainty quantification performance. The recognition inferiority can be attributed to the fact that contrastive loss only considers relations between a pair of samples, while triplet loss optimizes the embeddings of three samples at a time. \sysname\ (Quadruplet) marginally outperforms \sysname\, 
which we believe is due to the supervision provided by the additional negative sample.
Another observation
is that better recognition often leads to better uncertainty quantification. This is reasonable because when a network produces a higher recognition performance, the mean of the non-deterministic embedding is closer to the underlying embedding center and it is thus easier for the network to fit a distribution.

\subsection{Qualitative Analysis}
Fig. \ref{fig_query} presents query samples which \sysname\ assigns with different uncertainty levels. Our first observation is that query samples with large scene views tend to be assessed as certain by \sysname. We hypothesize that the building exteriors enrich the place characteristics and thus make a query be easily recognized. This trend can also be witnessed when we evaluate the pair uncertainty (i.e., the covariance of the query and its nearest neighbor as introduced by \cite{warburg2021bayesian}) shown in Fig. \ref{fig_pair}. We found that large pair uncertainty occurs when the query and the database sample have limited scene views. 

Secondly, we found that high-uncertainty query samples tend to have high image contrasts. To investigate the relation between image contrast and the predicted uncertainty, we first measure the image contrast by Histogram Spread (HS)\cite{tripathi2011performance}, where a high-contrast image would have a flat histogram and thus a high HS. The average HS of query samples with different uncertainty levels is further presented in Fig. \ref{fig_query}, where it is shown that query images with lower uncertainty levels generally have higher HS, i.e., higher image contrast. 
Interestingly, high contrast images were also found to yield better hand-crafted feature extraction performance\cite{zhao2010applying}. 


In Table. \ref{performance_table} we already demonstrate that uncertainty-aware training gives rise to superior recognition performance (i.e., \sysname\ vs Standard Triplet). We further follow the test convention of a similar task of face image quality verification \cite{grother2007performance, grother2019face}, and show the recognition correctness boost in Fig. \ref{fig_cleaning_query} when most uncertain queries are removed - mimicking the scenario when a place recognition system denies making a prediction for a given image due to uncertainty consideration and hands over the recognition to the human operator. The correctness ratio refers to the ratio of correctly matched query and top1 candidate pairs to all pairs. As we can see, \sysname\ shows a consistent improvement as more uncertain query samples are removed and obtains the highest recognition performance boost when 90\% most uncertain query samples are removed. These results imply that the place recognition performance can benefit from the estimated uncertainty during the inference phase, and \sysname\ yields the best quality uncertainty estimation.


%% file: sections/6_conclusion.tex
\section{Conclusion}

This paper proposes a self-teaching framework \sysname\ to estimate the per-exemplar uncertainty in a place recognition system. The experimental results show that (1) \sysname\ improves recognition performance by uncertainty-aware training, (2) \sysname\ estimates better-calibrated uncertainty than SOTA methods, and (3) \sysname\ is flexible to accommodate different loss functions widely used in place recognition. 
Furthermore, our experimental results suggest that to reduce retrieval uncertainty and improve recognition performance, a real-world place recognition system favors the capture of large scene views and high contrast images during both the inference and the database building phases.